\documentclass{article}

\usepackage{arxiv}

\usepackage[pagebackref=true]{hyperref}

\usepackage{graphicx}
\usepackage{url}
\usepackage{float}
\usepackage{subcaption}
\usepackage{tikz}
\usepackage{setspace}

\newcommand{\eqy}{\hat{y}}
\newcommand{\eqR}{\hat{R}}

\usepackage[dvipsnames]{xcolor}

\usepackage{xcolor}  % Needed for colours - load before mdframed

\usepackage{amssymb}
\usepackage{amsmath}
\usepackage{mathdots}

\usepackage{mathtools}

\usepackage{tensor}

\usepackage{urwchancal}
\DeclareFontFamily{OT1}{pzc}{}
\DeclareFontShape{OT1}{pzc}{m}{it}{<-> s * [1.10] pzcmi7t}{}
\DeclareMathAlphabet{\mathpzc}{OT1}{pzc}{m}{it}

\usepackage{graphicx}
\usepackage{ifpdf}
\ifpdf
\usepackage{epstopdf}
\PrependGraphicsExtensions{.svg}
\fi
\newtheorem{theorem}{Theorem}[section]

\newtheorem{proposition}[theorem]{Proposition}

\providecommand{\R}{\mathbb{R}}
\providecommand{\SL}{\mathbf{SL}}

\providecommand{\SE}{\mathbf{SE}}
\providecommand{\grpG}{\mathbf{G}}

\providecommand{\gothg}{\mathfrak{g}}

\providecommand{\se}{\mathfrak{se}}

\providecommand{\calM}{\mathcal{M}}

\providecommand{\torSE}{\mathcal{SE}}

\providecommand{\vecL}{\mathbb{L}}

\providecommand{\GP}{\mathbf{N}} % Gaussian noise process.
\DeclareMathOperator{\Ad}{Ad}
\DeclareMathOperator{\ad}{ad}

\providecommand{\id}{\mathrm{id}} % identity map
\providecommand{\td}{\mathrm{d}}
\providecommand{\tD}{\mathrm{D}}

\providecommand{\ddt}{\frac{\td}{\td t}}

\providecommand{\mr}[1]{\mathring{#1}} % reference element.

\providecommand{\ob}[1]{\overline{#1}} % homogeneous vector
\usepackage{accents}
\usepackage{mathtools}
\makeatletter
\providecommand{\scirc}{%
    \hbox{\fontfamily{\rmdefault}\fontsize{0.4\dimexpr(\f@size pt)}{0}\selectfont{\raisebox{-0.52ex}[0ex][-0.52ex]{$\circ$}}}}
\makeatother
\makeatletter
\providecommand{\ucirc}{%
    \hbox{\fontfamily{\rmdefault}\fontsize{0.4\dimexpr(\f@size pt)}{0}\selectfont{\raisebox{0.0ex}[0ex][-0.52ex]{$\circ$}}}}

\makeatother

\mathchardef\mhyphen="2D
\usepackage{stfloats}

\newcommand{\papercitation}{% add citation to published paper in text
This work was supported by the Australian Research Council under the Discovery Project DP250100112.
\emph{Accepted for presentation at IFAC World Congress}. 
}
\newcommand{\publicationdetails}{
    \copyrightNoticeIFACAccepted{2026} % Use for IEEE papers
    \papercitation
}
\newcommand{\publicationversion}
{Author accepted version}

\begin{document}

\title{Equivariant Filter for High Performance Image Tracking using an Event Camera}

\headertitle{
    Equivariant Filter for High Performance Image Tracking using an Event Camera
    }  
    \author{
Angus Apps
\\
    Systems Theory and Robotics Group \\
    School of Engineering \\
	Australian National University \\
    Canberra, Australia \\
    \texttt{Angus.Apps@anu.edu.au} \\
    \And    
Yixiao Ge
\\
    Systems Theory and Robotics Group \\
    School of Engineering \\
	Australian National University \\
    Canberra, Australia \\
    \texttt{Yixiao.Ge@anu.edu.au} \\
    \And    
Timothy~L. Molloy
\\
    School of Engineering \\
	Australian National University \\
    Canberra, Australia \\
	\texttt{Timothy.Molloy@anu.edu.au} \\
    \And	
Robert Mahony
\\
    Systems Theory and Robotics Group \\
    School of Engineering \\
	Australian National University \\
    Canberra, Australia \\
	\texttt{Robert.Mahony@anu.edu.au} \\
}

\maketitle

%====================================================%
%=====                 Abstract                 =====%
%====================================================%
\begin{abstract}                % Abstract of 50--100 words
Image tracking is the problem of estimating the transformation that relates a moving image of a scene to an original reference image. 
The problem is important in control of autonomous vehicles or robots, where the image encodes information about the motion of the camera or environment, as well as in pure  computer vision applications. 
In this paper, we present an equivariant filter design for high performance tracking of planar image transformations using an event camera.
The design exploits the Asynchronous Event Blob (AEB) tracker \cite{wang2024asynchronous} to extract feature-position measurements from the raw event stream, and an equivariant filter to compute an affine image translation and rotation using the special Euclidean group symmetry.
The equivariant filter incorporates an equivalent-measurement update step that de-correlates the (highly temporally correlated) feature-position measurements provided by the AEB tracker.
We evaluate the design experimentally using two datasets involving general and fast rotational motion. 
We benchmark results against direct optimisation (estimating the relative transformation from the raw blob tracks), and a covariance intersection approach for overcoming data correlation. 
Our design provides smooth image tracking for features moving up to 7000 pixels per second on the image plane.
\end{abstract}

%====================================================%
%=====               Introduction               =====%
%====================================================%
\section{Introduction}

Image tracking is the problem of estimating the transformation that relates a moving image of a scene to an original reference image. 
Image tracking is widely used in many computer vision applications, such as perspective correction \cite{corke2011robotics} and image stabilisation \cite{hua2019feature}, and has proven valuable in many robotics applications, such as visual odometry \cite{scaramuzza2008appearance} and vision-based control \cite{benhimane2007homography, lai2023homography}.
Classical computer vision techniques exploit the algebraic relationship between sets of matching feature points to compute the corresponding image transformation \cite{hartley2003multiple}.
With sufficient matching points, algebraic algorithms can compute quality estimates of the transformation between consecutive image frames.
However, these estimates are computed independently for each pair of images and do not exploit the temporal evolution of the image transformation caused by real-world camera motion.
To address this issue, authors in the systems and control community have developed non-linear observers for image transformation estimation. 
In the last decade, deterministic observers for homography tracking were developed  \cite{hamel2011homography,mahony2012nonlinear} that provide robust performance with strong stability guarantees. 
Recent years have seen the development of stochastic filters for homography tracking. 
\cite{bernal2023bayesian} proposed an interacting multiple model filter that runs parallel iterated extended Kalman filters with different process noise to manage model mismatch.
\cite{bouazza2023equivariant} proposed an equivariant filter to estimate homography and structure parameters for a moving image sequence. 
This work allowed incorporation of inertial measurement unit inputs to predict the image motion rather than direct estimation of the infinitesimal homography velocity. 
Stochastic filtering has been applied effectively in other image processing applications, such as recursive structure and motion estimation in 3D vision \cite{ma2004invitation} and vision-based state estimation using infinite-dimensional measurements \cite{varley2025}.

A fundamental challenge for existing image tracking algorithms is feature tracking, particularly in highly dynamic scenarios where high optic flow causes large feature displacement between images.
Event cameras introduce a new sensing modality that provides a significant advantage for tracking in highly dynamic environments \cite{gallego2020event}.
Event-based feature tracking has been approached using both model- and learning-based approaches.
While learning-based approaches are effective for tracking general objects \cite{hamann2025etap}, they require temporal accumulation or discretisation, sacrificing the low-latency advantage of event-driven data. 
For high performance applications where simple features are tracked, model-based approaches have been shown to be preferable \cite{delbruck2008frame,wang2024asynchronous}.
\cite{wang2024asynchronous} propose the Asynchronous Event Blob (AEB) tracker that models features as spatial Gaussian distributions whose parameters are estimated by an extended Kalman filter.
The AEB tracker has shown robust tracking at over 11,000 pixels per second.
Although the AEB tracker offers significant advantages, the output tracks generated by the AEB tracker are highly correlated due to the filtering process and cannot be treated as independent measurements in a subsequent image tracking filter.  
Managing correlated measurements has been explored in multi-sensor fusion applications, particularly in track-to-track fusion algorithms \cite{chong2000architectures,chang2002optimal}.
These situations occur when there are multiple different nodes or sensors providing an estimate of the same state variable that are fused to generate a central estimate.
Here, the correlation is often unknown and can cause filter instability if ignored \cite{julier2017general}.
\cite{frenkel1994flexible} proposed an equivalent-measurement framework that generates a measurement that would have generated the same estimate evolution, but remains temporally independent.
This approach has been used in applications such as bias estimation \cite{okello2004joint,taghavi2016practical}. 

In this paper, we propose an Equivariant Filter (EqF) to estimate a planar image transformation using event cameras. 
We develop a cascaded filter design whereby the AEB tracker processes the raw event stream to output feature positions.
These tracks are then used asynchronously in an image tracking filter.
However, since the AEB tracker outputs have high temporal correlation, they cannot be directly used as measurements in the image tracking filter.
To overcome this, we use the equivalent measurement framework to provide uncorrelated track estimates that can be used directly in a second stage EqF for image tracking. 
To demonstrate our approach and simplify the exposition we focus on planar transformations corresponding to $\SE(2)$ image transformations.
The extension of these ideas to full image homography estimation on $\SL(3)$ is algebraically more complicated and is left for future work, although the authors expect the results to be similar. 
We evaluate our algorithm experimentally on two datasets and compare to both a covariance intersection fusion algorithm, and a raw ``event-by-event'' optimisation algorithm.
Our algorithm provides significantly smoother and more stable estimates than the alternate algorithms for features moving up to 7000 pixels per second.

%====================================================%
%=====               Preliminaries              =====%
%====================================================%
\section{Preliminaries} \label{sec:preliminaries}

Let $\grpG$ be a general Lie group with dimension $n$, associated with the Lie algebra $\gothg$.
Let $\id$ denote the identity element of $\grpG$.
For any given $X,Y\in\grpG$, the left and right translations are denoted by $\textrm{L}_X, \textrm{R}_X : \grpG \to \grpG$, and are defined by 
\begin{align*}
    \textrm{L}_X(Y):=XY, \quad \textrm{R}_X(Y):=YX.
\end{align*}
The Lie algebra $\gothg$ is isomorphic to a vector space $\R^n$ with the same dimension.
We use wedge $(\cdot)^\wedge:\R^n\rightarrow\gothg$ and vee $(\cdot)^\vee:\gothg\rightarrow\R^n$ operators to map between the Lie algebra and the vector space.
The Adjoint map for the group $\grpG$, $\Ad_X:{\gothg}\to{\gothg}$ is defined by
\begin{align*}
        \Ad_{X}[{{u}^{\wedge}}] = \tD \textrm{L}_{X} \circ\tD \textrm{R}_{X^{-1}}\left[{u}^{\wedge}\right] ,
\end{align*}
for every $X \in \grpG$ and ${{u}^{\wedge} \in \gothg}$, where $\tD \textrm{L}_{X}$, and $\tD \textrm{R}_{X}$ denote the differentials of the left and right translations, respectively.
The adjoint map for the Lie algebra $\ad_{{u}^\wedge}: {\gothg}\to{\gothg}$ is given by
\begin{equation*}
    \ad_{{u}^\wedge}{{v}^{\wedge}} = \left[{u}^{\wedge}, {v}^{\wedge}\right] ,
\end{equation*}
and is equivalent to the Lie bracket.

A (right) group action of a Lie group $\grpG$ on a manifold $\calM$ is a smooth map $\phi:\grpG\times\calM\rightarrow\calM$ that satisfies
\begin{align*}
    \phi(X,\phi(Y,\xi))=\phi(YX,\xi) && \textrm{and} && \phi(\id,\xi)=\xi
\end{align*}
for all $X,Y \in\grpG$ and $\xi\in \calM$.
It induces the partial maps $\phi_X:\calM\rightarrow\calM$ and $\phi_\xi:\grpG\rightarrow\calM$ which are defined by $\phi_X(\xi):=\phi(X,\xi) =: \phi_\xi(X)$.

%====================================================%
%=====            Problem Formulation           =====%
%====================================================%
\section{System Model} \label{sec:system_model}

Consider an event camera translating and rotating over a planar scene
perpendicular to the optical axis of the camera. 
Since we are only considering planar motion, we can rescale the image so that the image coordinates correspond to the Euclidean coordinates, saving considerable complexity in the formulation. 
Choose an arbitrary 2D inertial reference frame $\{0\}$ and let $\{B\}$ denote the camera's 2D body-fixed frame.
That is, the reference image has coordinates $\{0\}$ and the body-frame image has coordinates $\{B\}$. 
We will assume that the event camera tracks a collection of blob-like features on the planar scene. 
Let $\mr{p}^i \in \{0\}$ for $i \in \{1, \ldots, n\}$ denote the stationary locations of the feature points in the reference image. 
The position of the blob in the body-frame image at time-index $t$ is denoted $p^i_t \in \{B\}$. 

We would like to estimate the image transformation from $\{B\}$ to $\{0\}$ and its linear and angular velocity in $\{B\}$. 
Since only planar motion is considered, then the associated image transformation can be represented by a homogeneous transformation matrix $P \in \SE(2)$ in the special Euclidean group on two dimensions. 
That is, $P \in \SE(2)$ represents the special Euclidean transformation from $\{0\}$ to $\{B\}$ (at an implicit time-index $t$ that will be clear from context). 
It follows that the true position of the $i$th feature in the body frame $\{B\}$ is given by
\begin{align}
    \ob{p}^{i}_t = \ob{h_i(P)} =  P^{-1} \ob{\mr{p}}^i, 
\end{align}
where the overbar indicates homogeneous coordinates.
The function $h_i(P)$ (no overbar) returns $p^i_t \in \R^{2}$.
The measurement model used for image tracking is 
\begin{align}
    y^{i}_t = h_i(P) + \mu_t, \quad \mu_t \sim \GP(0,R_t),  \label{eq:base_measurement}
\end{align}
which is a measurement of the $i$th feature-position in the body frame $\{B\}$ at time $t$. 
Note that we do not actually have an independently distributed measurement $y^i_t$ available and we will need to construct this from the output of the AEB tracker as discussed in Section \ref{sec:update_equivalent_measurement}. 

We will use a constant velocity model for the proposed image tracking algorithm. 
The state space is identified as 
\begin{align}
    \calM:= \torSE(2)\times\se(2),
\end{align}
where $\torSE(2)$ is the $\SE(2)$-torsor with elements representing the camera position and rotation on the plane.
The $\se(2)$ part of the state is the angular and linear velocity of the camera in $\{B\}$ corresponding to the angular and linear velocity components of the 2D special Euclidean transformation.
The element $P \in \torSE(2)$ represents the camera orientation and position on the plane expressed as a homogeneous transformation matrix while the camera's linear and angular velocities are denoted by $V\in\se(2)$.

The deterministic system dynamics are given by 
\begin{align}
    \dot{P} = P(V + {U_{1}}^{\wedge}), \qquad\dot{V} = {U_{2}}^\wedge.\label{eq:dynamics}
\end{align}
The first input signal $U_1\in\R^3$ is a virtual input that is included to provide structure for the symmetry operation \cite{ng2020equivariant} 
and will be set to zero in the algorithm. 
The second input signal $U_2\in\R^3$ is the acceleration input and is set to zero since we assume constant velocity. 
That is, the input velocity of the system in the filter implementation is $(U_1, U_2) = (0_{3\times1}, 0_{3\times1})\in\vecL$, where $\vecL:= \R^3 \times \R^3$ is the input space.

%====================================================%
%=====                 Algorithm                =====%
%====================================================%
\section{Equivariant filter design} \label{sec:algorithm}

In this section, we develop the proposed Equivariant Filter (EqF) design for the image tracking problem according to the procedure of \cite{van2022equivariant}.

%========================================================================================================%
\subsection{Symmetry of the system} \label{sec:symmetry}

The system symmetry follows \cite{ng2020equivariant}.
Let $Z:=(P, V)\in\calM$ denote elements of the state space. 
Define 
\begin{align}
\dot{Z} = \ddt(P, V) = f_{(U_1, U_2)}(P, V) = f_{U}(Z)  
\end{align}
with $U = (U_1, U_2) \in \vecL$ as the compact form of 
\eqref{eq:dynamics}. 

Define a symmetry Lie group $\grpG:=\SE(2)\ltimes\se(2)$ with elements written $X = (A, a) \in \grpG$ \cite{ng2020equivariant}. 
The group product, identity and inverse are given by
\begin{align*}
    (A_{1}, a_{1})\cdot(A_{2}, a_{2}) = (A_{1}A_{2}, a_{1} + \Ad_{A_{1}}a_{2}),
\end{align*}
\begin{align*}
    \id = (I_3, 0_3),\qquad (A_{1}, a_{1})^{-1} = (A^{-1}, -\Ad_{A^{-1}}a).
\end{align*}
The symmetry actions and system lift required for the EqF are given in the following propositions proved in (\cite{ng2020equivariant}, Section IV. and V., respectively).

\begin{proposition}[State action]
    The group action $\phi \colon \grpG \times \calM \rightarrow \calM$ defined by 
    \begin{align}\label{eq:state_action}
        \phi((A,a), (P_{Z},V_{Z})) = (P_{Z}A, \Ad_{A^{-1}}(V_{Z}-a)) 
    \end{align}
    is a transitive right group action.
\end{proposition}

\begin{proposition}[Input action]
    Define $\psi \colon \grpG \times \vecL \rightarrow \vecL$ as 
    \begin{align}\label{eq:input_action}
    \psi((A,a),(U_{1}, U_{2})) = (\Ad_{A^{-1}}({U_{1}}^\wedge + a), \Ad_{A^{-1}}{U_{2}}^\wedge).
    \end{align}
    Then $\psi$ is a right group action of $\grpG$ on $\vecL$.
\end{proposition}

\begin{proposition}[System lift]
    Define the map $\Lambda:\calM\times\vecL\rightarrow \gothg$ by
    \begin{align}\label{eq:lift}
    \Lambda\left((P_{Z},V_{Z}), (U_{1}, U_{2})\right) &= \left(V_{Z} + {U_{1}}^\wedge, \ad_{V_{Z}}{U_{1}}^\wedge-{U_{2}}^\wedge \right).
    \end{align}
    % \begin{flalign}\label{eq:lift}
    % \Lambda\left((P_{Z},V_{Z}), (U_{1}, U_{2})\right) &= \left(V_{Z} + {U_{1}}^\wedge, \ad_{V_{Z}}{U_{1}}^\wedge-{U_{2}}^\wedge \right). & \raisetag{\baselineskip}
    % \end{flalign}
    Then $\Lambda$ is a lift for the system in \eqref{eq:dynamics} with respect to the symmetry group; that is, it satisfies the lift condition
    \begin{align*}
        \left.\mathrm{D}_E\right|_{\mathrm{id}} \phi_{\left(P, V\right)}(E) \left[\Lambda\left(\left(P, V\right),(U_1, U_2)\right)\right]=f_{(U_1, U_2)}\left(P, V\right).
    \end{align*}
\end{proposition}

%========================================================================================================%
\subsection{Equivariant filter dynamics}

Choose an origin $\mr{Z}\in\calM$ to be $\mr{Z}:=(\mr{P}, \mr{V}) = (I_3, 0_3)$.
Define the local coordinate chart $\vartheta:\mathcal{U}_{\mr{Z}}\subset\calM\rightarrow \R^6$ to be
\begin{align}\label{eq:local_coordinate}
    \vartheta:=\left( \phi_{\mr{Z}} \cdot \exp_\grpG \right)^{-1} = \log_\grpG \cdot \phi_{\mr{Z}}^{-1},
\end{align}
on a neighborhood $\mathcal{U}_{\mr{Z}}$ around the origin such that the map is bijective.
The logarithm $\log_\grpG$ in \eqref{eq:local_coordinate} is computed using the matrix realisation of the semi-direct product Lie group.

Let $Z = (P, V)\in\calM$ denote the true state of the system. 
Define the observer state to be $\hat{X} =(\hat{A}, \hat{a})\in\grpG$ with covariance $\Sigma\in\mathbb{S}_+(6)$, where $\mathbb{S}_+(n)$ denotes the set of $n \times n$ symmetric positive definite matrices.
The covariance matrix $\Sigma$ is a positive definite operator on the vector representation of the direct product Lie-algebra $\gothg \times \gothg \equiv \R^6$. 
The estimate of the state is obtained by applying the group action to the origin $\hat{Z} = \phi_{\hat{X}}\mr{Z}$.
Let $y^i_t\in \R^2$ denote the output track coordinates of a the $i$th tracked point provided by the AEB front end tracker. 

The equivariant error is given by $e = \phi(\hat{X}^{-1}, Z)$, and the filter dynamics are derived by linearising the dynamics of the local state error $\varepsilon=\vartheta(e)$.
Define $A_t$, $B_t$ and $C_t$ to be the Jacobian matrices of the EqF, defined in \cite{van2022equivariant}. 
The input $U_1 = 0$ is set to zero to enforce the system kinematics.
We use a constant velocity model for the dynamics and set $U_2 \equiv 0$. 
The filter is designed to run predict and update synchronously.
The discrete filter update equations for a single measurement are given by
\begin{align}
    & \hat{X}_{t\mid t-1} \!=\! \hat{X}_{t-1\mid t-1} \exp_\grpG( \delta t \, \Lambda (\phi(\hat{X}_{t-1\mid t-1}, \mr{Z}), (0, 0))) \\
    & \Sigma_{t\mid t-1} = \exp(A_{t} \delta t) \, \Sigma_{t-1\mid t-1} \, \exp(A_{t} \delta t)^{\top} + \delta t B_{t} Q_{t} B_{t}^{\top} \\
     & \Delta=\tD_E\big|_{\mathrm{id}} \phi_{\mr{Z}}(E)^{\dagger} \, \tD \vartheta^{-1} \Sigma_{t\mid t-1} {C_t}^\top (C_t \Sigma_{t|t-1} {C_t}^\top + R_t)^{-1} (y^i_t-h_i(\phi(\hat{X}_{t\mid t-1}, \mr{Z}))) \label{eq:eqf_delta} \\
    & \hat{X}_{t\mid t} = \exp_\grpG(\Delta) \hat{X}_{t\mid t-1} \\
    & \Sigma_{t\mid t} = (I-\Sigma_{t\mid t-1}{C_t}^\top(C_t \Sigma_{t|t-1} {C_t}^\top + R_t)^{-1}C_t)\Sigma_{t\mid t-1}.\label{eq:eqf_cov_update} 
\end{align}

The matrices $Q_t\in\mathbb{S}_+(6)$ and $R_t\in\mathbb{S}_+(2)$ are the input and output gain matrices, $\Delta\in\gothg$ is the correction term, and $\delta t$ is the time since the last EqF update.

%====================================================%
%=====          Equivalent Measurement          =====%
%====================================================%
\section{Equivalent Measurement Update} \label{sec:update_equivalent_measurement}

As discussed, the outputs of event-based AEB feature trackers are used as point measurements for image tracking.
These track estimates have strong temporal correlation as they are generated by a filter that correlates the full measurement history to generate the present estimate of the state. 
To compensate for this in the EqF, we propose to use the equivalent measurement formulation introduced by \cite{frenkel1994flexible}.
% \blue{
An advantage of this approach is that it also allows one to update the EqF at a slower rate than the AEB tracker, reducing computational load when required.
% }

\begin{figure}[t]
    \centering
    % \resizebox{0.5\columnwidth}{!}{\input{Figures/equivalent_measurement_figure}}
    \resizebox{0.5\columnwidth}{!}{\begin{tikzpicture}[>=stealth, thick]
    \usetikzlibrary{arrows.meta}
    \coordinate (A) at  (0,0);
    \coordinate (B1) at (0.25,0.3);
    \coordinate (B2) at (0.25,-0.3);
    \coordinate (C1) at (7,0.3);
    \coordinate (C2) at (7,-0.3);
    \coordinate (D)  at (7,0);

    % Styles
    \tikzstyle{traj}=[smooth]
    \tikzstyle{dline}=[dashed]

    %----- Dashed Lines -----%
    \draw[dline] (0,-0.5) -- (0,0.6);
    \node[above] at (0,-1) {$k-m$};

    % Right dashed vertical (k)
    \draw[dline] (7,-0.5) -- (7,-0.25);
    \draw[dline] (7,0.3) -- (7,0.6);
    \node[above] at (7,-1) {$k$};

    %----- Prediction Curve -----%
    % Predict arrow
    \draw[->, >= {Stealth[scale=1.25]}, align=center] (B1) -- (C1) node[midway, above] {Predict Only: $(\hat{p}^{i}_{k|k-m},\Sigma^{i}_{k|k-m})$};
    \draw    (A) to[out=90,in=-180] (B1);
    \draw[dotted]    (C1) to[out=90,in=90] (D);

    % Predicted point on the right
    \fill (A) circle (1.5pt);
    \node [left, align=center] at (A) {$\hat{p}^{i}_{k-m \mid k-m}$ \\ $\Sigma^{i}_{k-m \mid k-m}$};    % \draw[dotted, thin]    (E1) to[out=0,in=90] (A);
    % \draw[->, >= {Stealth[scale=1.25]}, thin]   (E2) to[out=0,in=-90] (A);

    %----- Updated Curve -----%
    \draw[->, >= {Stealth[scale=1.25]}, align=center] (B2) -- (C2) node[midway, below] {Predict and Update: $(\hat{p}^{i}_{k|k}, \Sigma^{i}_{k|k})$};
    \draw (A) to[out=-90,in=180] (B2);
    \draw (C2) to[out=-90,in=-90] (D);

    % Predicted point on the right
    \fill (D) circle (1.5pt);
    \node [right, align=center] at (D) {$\hat{p}^{i}_{k \mid k}$ \\ $\Sigma^{i}_{k \mid k}$};
    \fill[red] (C1) circle (1.5pt);
    \fill[red] (C2) circle (1.5pt);

\end{tikzpicture}   }
    \caption{
    Equivalent measurement formulation for the $i$th feature.  
    From index $k-m$, compute the predicted and updated AEB states up to index $k$. The equivalent measurement is computed from these by \eqref{eq:equivalent_measurement_value}--\eqref{eq:equivalent_measurement_covariance}.
    }
    \label{fig:equivalent_measurement_diagram}
\end{figure}
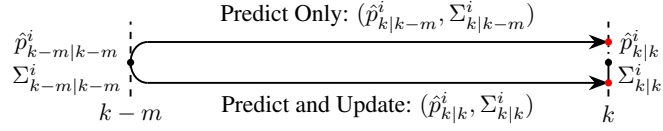

Let $t$ denote the index of the EqF filter and let $k$ denote the index of the output of the $i$th AEB tracker. 
Let $t_k^i$ denote the EqF filter index at a time coincident with the $i$th AEB tracker estimate with index $k$. 
Consider the EqF update at a time $t_k^i$ based on the historic track data $\{k - m + 1, \ldots, k\}$, where $m \in \{1, 2, \ldots\}$ denotes the number of events accumulated between EqF updates. 
Typically, we will use $m = 1$, using information from just the latest event, and run the EqF at the same rate as the AEB tracker updates.
However, by accumulating data we can ensure that the equivalent measurement remains well defined and reduce computational load if required. 
Let $(\hat{p}^{i}_{k \mid k}, \Sigma^{i}_{k \mid k})$ denote the AEB filter estimate, including covariance estimate, at index $k$.
Note that the notation $\Sigma$ is overloaded to represent both the covariance estimate from the AEB filter and EqF, and they differ by the use of $k$ and $t$ in the subscript, respectively.
We modify the AEB tracker to generate the accumulated prediction filter-state estimate $(\hat{p}^{i}_{k \mid {k-m}}, \Sigma^{i}_{k \mid {k-m}})$ from ${k-m}$ to $k$. 
If $m = 1$ then this is exactly the standard prediction provided by a Kalman filter. 
If $m > 1$ then this is provided by iterating the prediction equations of the AEB filter without update for indexes ${k-m}$ to $k$ (Fig.~\ref{fig:equivalent_measurement_diagram}). 
The equivalent measurement formulation considers an independently distributed virtual measurement 
\begin{align}
    \eqy^{i}_{k\mid k-m} &= p^{i}_{k} + \eta_{k\mid k-m}, \quad \eta_{k\mid k-m} \sim \GP (0, \eqR^{i}_{k\mid {k-m}}) \label{eq:equivalent_measurent}
\end{align}
that would have generated the updated-state $(\hat{p}^{i}_{k \mid k}, \Sigma^{i}_{k \mid k})$ from the predicted-state $(\hat{p}^{i}_{k \mid {k-m}}, \Sigma^{i}_{k \mid {k-m}})$ for a single measurement update . 
Here $p^{i}_{k}$ is the true feature-position and $\eqR^{i}_{k\mid {k-m}}$ is a measurement covariance associated with a (virtual) zero mean independent Gaussian noise process $\eta_{k\mid k-m}$. 
The measurement $\eqy^{i}_{k\mid k-m}$ and its noise covariance $\eqR^{i}_{k\mid {k-m}}$ can be computed from the state estimates by implementing the standard Kalman filter update equations in reverse \cite{frenkel1994flexible, welch1995introduction},
\begin{align}
    \eqy^{i}_{k\mid k-m} &= \hat{p}^{i}_{k \mid {k-m}} + \Sigma^{i}_{k \mid {k-m}} \left(\Sigma^{i}_{k \mid {k-m}} - \Sigma^{i}_{k \mid k}\right)^{-1} \left(\hat{p}^{i}_{k \mid k} - \hat{p}^{i}_{k \mid {k-m}}\right) \label{eq:equivalent_measurement_value} \\
    \eqR^{i}_{k\mid {k-m}} &= \Sigma^{i}_{k \mid {k-m}} \left(\Sigma^{i}_{k \mid {k-m}} - \Sigma^{i}_{k \mid k}\right)^{-1} \Sigma^{i}_{k \mid {k-m}} - \Sigma^{i}_{k \mid {k-m}} \label{eq:equivalent_measurement_covariance}.
\end{align}
The equivalent measurement $\eqy^{i}_{k\mid k-m}$ is an independently distributed function of the true track position $p^i_k$ which in turn is a deterministic function of the image transformation state. 
We set $y^{i}_{t_k} = \eqy^{i}_{k\mid k-m}$ in \eqref{eq:eqf_delta} and $R^i_{t_k} = \eqR^i_{k \mid k-m}$ in \eqref{eq:eqf_cov_update} to implement the image tracking EqF. 

It is important to note that the inverse of the change in covariance $(\Sigma^i_{k \mid {k-m}} - \Sigma^i_{k\mid k})$ in \eqref{eq:equivalent_measurement_value}--\eqref{eq:equivalent_measurement_covariance} may be ill conditioned. 
This is particularly the case for the output of the AEB tracker where each event (update) introduces very little information.
As the difference approaches zero, numerical precision can introduce instability in the matrix inverse. 
In such situations, one must choose $m > 1$ to increase the accumulation window until there is sufficient information in the equivalent measurement to make \eqref{eq:equivalent_measurement_value} and \eqref{eq:equivalent_measurement_covariance} well defined. 
The value of $m$ can be chosen dynamically by testing $(\Sigma^i_{k \mid {k-m}} - \Sigma^i_{k\mid k})$ at each index to check that the inverse is well conditioned. 

%====================================================%
%=====            Experimental Setup            =====%
%====================================================%
\section{Experimental Setup} \label{sec:experiment_setup}

In this section, we outline the experimental setup used to evaluate the algorithm proposed in Section \ref{sec:algorithm} and \ref{sec:update_equivalent_measurement}.

%========================================================================================================%
\subsection{Hardware Setup} \label{subsec:hardware_setup}

\begin{figure}[h]
    \centering
    %--- Row 1 ---%
    \begin{subfigure}{0.49\linewidth}
        \centering
        \includegraphics[height=50mm]{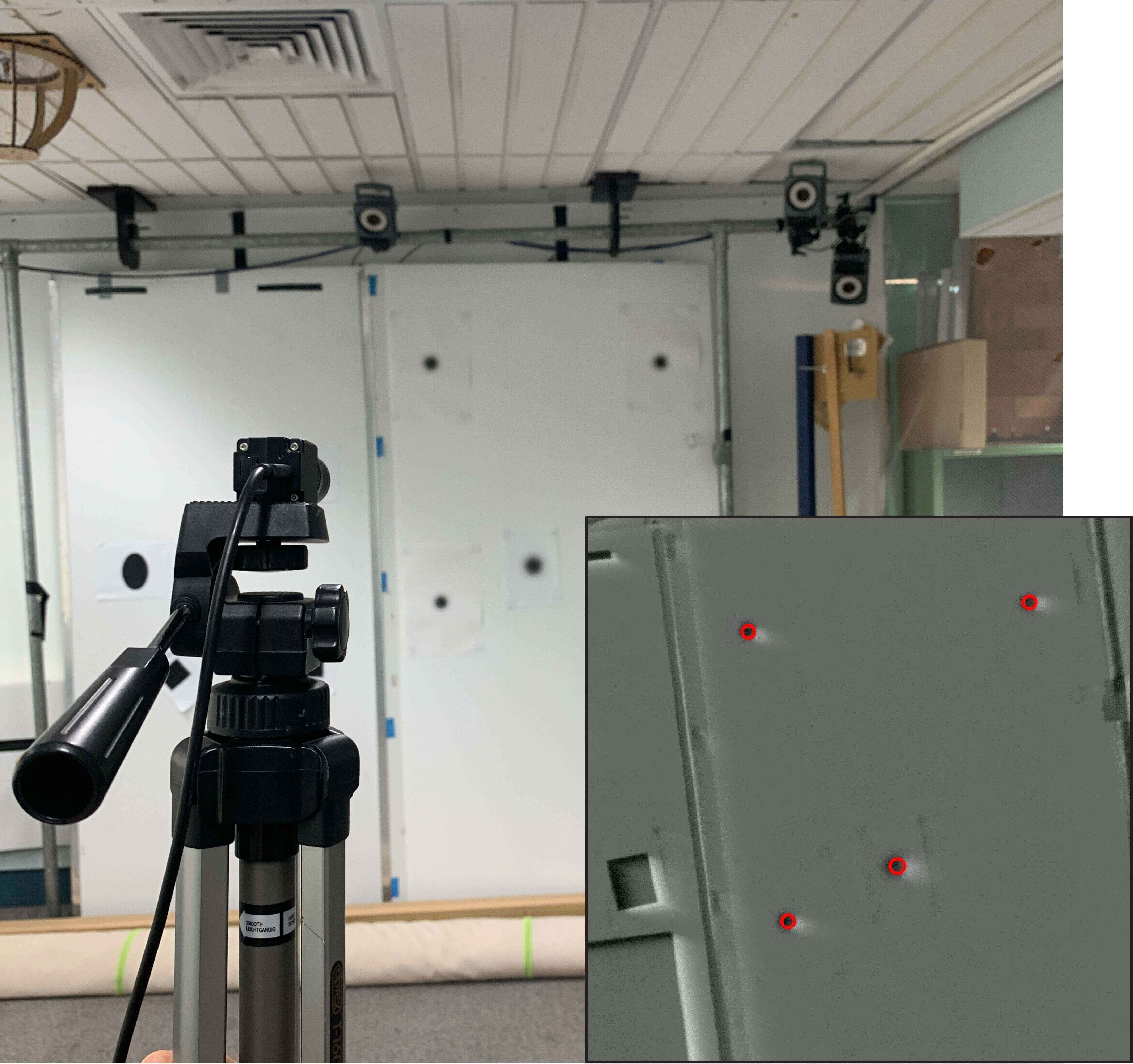}
        \caption{}
        \label{fig:general_exp_setup}
    \end{subfigure}
    % \hfill
    \begin{subfigure}{0.49\linewidth}
        \centering
        \includegraphics[height=50mm]{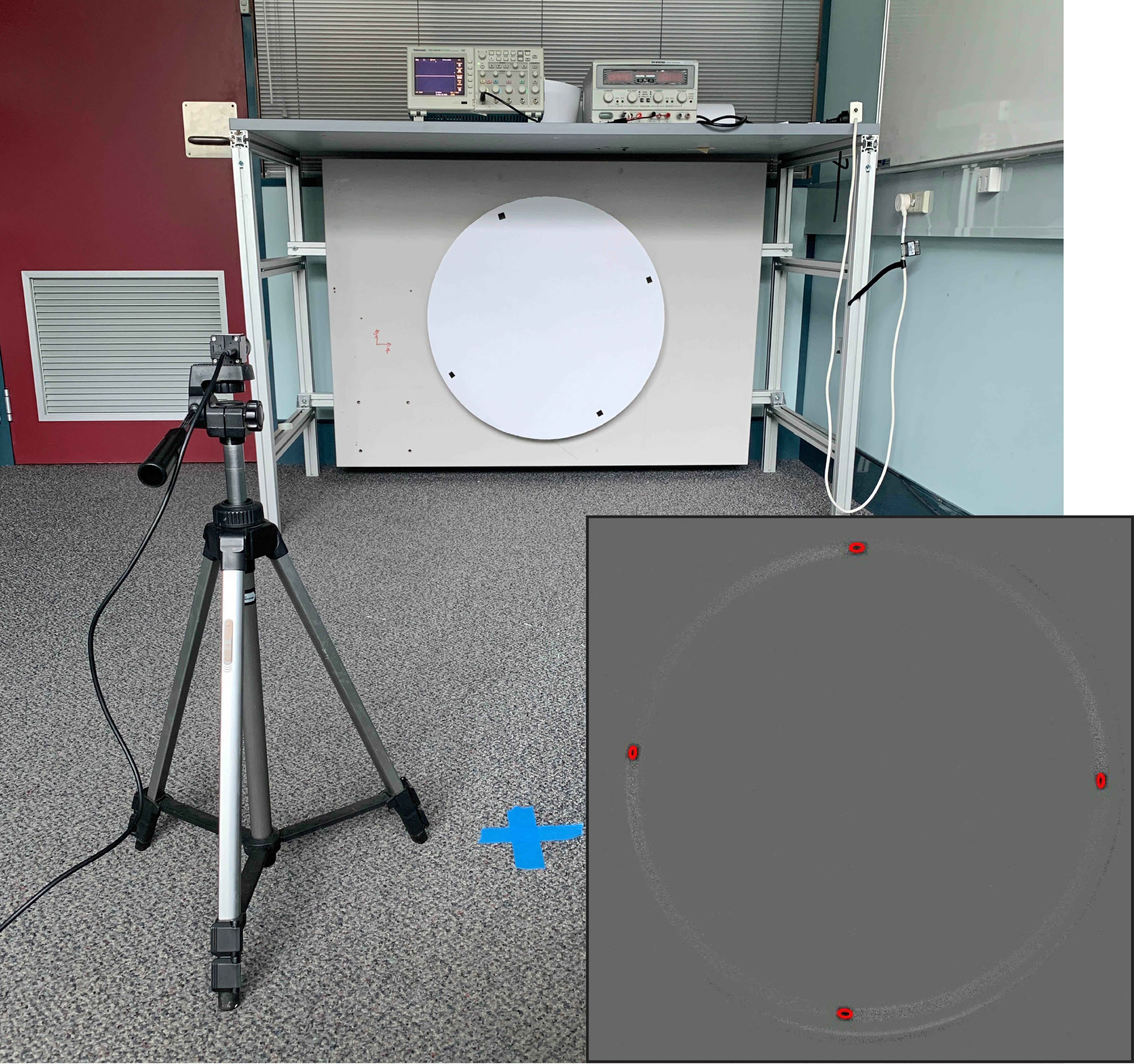}
        \caption{}
        \label{fig:spinning_exp_setup}
    \end{subfigure}
    \caption{
        Experimental setup for (\subref{fig:general_exp_setup}) general planar motion and (\subref{fig:spinning_exp_setup}) fast rotating motion. Inset images show the features (red) as observed from the AEB tracker. % \citep{wang2024asynchronous}.
    }
    \label{fig:experimental_setup}
\end{figure}

Two experiments were conducted with the Prophesee EVK4 event camera.
To evaluate the algorithm under general $\se(2)$ motion, we marked a planar surface with four feature points.
The camera was moved by hand in the parallel plane (translation and rotation), shown in Figure \ref{fig:general_exp_setup}.
Since this data was based on a hand-held camera, there will be noticeable variation from true planar motion, and this data provides an indication of the robustness of the algorithm. 
To evaluate our algorithm in a high-speed application, four feature points were placed on a spinning planar disc attached to a stepper motor, shown in Figure \ref{fig:spinning_exp_setup}. 
 In Section \ref{subsec:results_fast}, the disc speed was $\sim$4.4 revolutions per second, corresponding to feature velocities of $\sim$7000 pixels per second on the image plane.
Although this setup uses a fixed camera and moving feature points, it can equally be interpreted as fixed feature points and a moving camera. 
Estimating the $\torSE(2)$ transformation instantaneously requires a minimum of two pairs of corresponding points.
For four points the estimation problem is over-determined, providing robustness to noisy measurements.
The tracking algorithm will continue to function even if points are occluded as long as there is at least one point visible to generate events that lead to filter updates. 
The filter will drift from the true solution if there are less than two points visible for an extended period of time. 

%========================================================================================================%
\subsection{Feature Tracking using the AEB Tracker} \label{subsec:aeb_tracker}

Feature points were tracked using the AEB tracker as it has shown state of the art performance for high-speed tracking \cite{wang2024asynchronous}. 
Consider a feature that is stationary in the world that is being observed by a moving event camera.
A feature is considered an \emph{event blob} when the spatial distribution of events it generates on the image plane is well-approximated by a bivariate Gaussian distribution around a moving center.
The AEB tracker parameterises an event blob by its position (mean) and orientation with associated linear and angular velocities, and its shape defined by the principal correlations of the Gaussian.
The state of the AEB tracker is estimated asynchronously with each associated event using an extended Kalman filter.
For image tracking, the AEB tracker provides position information of features at high temporal resolution.

%========================================================================================================%
\subsection{Comparisons} \label{subsec:comparisons}

We compare our algorithm to two alternatives.
First, we implement an algebraic optimisation for the least squares solution of the $\torSE(2)$ transformation between measured points in $\{B\}$ and reference points in $\{0\}$ at each time step \cite{umeyama2002least}.
Zero-order hold is used to sample all AEB trackers at the same time. 
The velocity components are estimated using first order finite difference over $\torSE(2)$ estimates.
For numerical stability, velocity estimates are computed over larger timescales than the raw event rate to reduce the effects of per event noise. 
At each time step, the velocity is computed between the current $\torSE(2)$ estimate and the estimate five milliseconds ago.
The orientation used to transform to a body velocity is taken as the average orientation between the two time instances.

We also compare our algorithm to the same EqF but using covariance intersection for data fusion in the update step \cite{julier2017general}. 
Measurements are taken directly as the AEB position and associated covariance, and are fused with the predicted EqF state via
\begin{align}
    & \Sigma_{t\mid t} =\left( \alpha_{t} \Sigma_{t\mid t-1}^{-1} + (1-\alpha_{t}) C_{t}^{\top} \Sigma^{i^{-1}}_{k \mid k} C_{t}  \right)^{-1} \label{eq:covariance_intersection_sigma}\\
    & \Delta = \Sigma_{t \mid t} (1-\alpha_{t}) C_{t}^{\top} \Sigma^{i^{-1}}_{k \mid k} (\hat{p}^i_{k \mid k} -h_i(\phi(\hat{X}_{t\mid t-1}, \mr{\xi}))) \label{eq:covariance_intersection_delta}\\
    & \hat{X}_{t\mid t} = \exp_\grpG(\Delta) \hat{X}_{t\mid t-1}
\end{align}
where $(\hat{X}_{t\mid t-1}, \Sigma_{t\mid t-1})$ is the predicted EqF state and $(\hat{p}^{i}_{k \mid k}, \Sigma^{i}_{k \mid k})$ is the updated AEB tracker position.
The scalar $\alpha_{t}$ is a weighting parameter that is chosen to minimise the trace of the updated covariance matrix $\Sigma_{t \mid t}$.

%====================================================%
%=====                 Results                  =====%
%====================================================%
\section{Results and Discussion} \label{sec:results}

In this section, we present the results of the experiments.

%========================================================================================================%
\subsection{Image Tracking under General Motion} \label{subsec:results_general}

We first compare the performance of the three algorithms for general $\se(2)$ motion including rotation and translation (Fig.\ref{fig:general_exp_setup}).
The state estimates are shown in Figure \ref{fig:genera_estimate_EM_OPT}.
At macro-scale, all algorithms appear to perform comparably for the $\torSE(2)$ estimate. 
However, it is noticeable that significantly more noise is introduced for the optimisation and covariance intersection methods in the $\se(2)$ estimates.
The difference in performance becomes more apparent at a finer time scale (right column).
For the $\torSE(2)$ estimate in the zoomed plots, the mean values remain similar, however the optimisation appears noisier than the equivalent measurement (no covariance estimate is available). 
Covariance intersection appears to have more low frequency deviations and maintains a significantly more conservative estimate.
The velocity terms reveal a larger disparity in performance as optimisation shows large variance in estimates (150-250 pixels per second) over millisecond time horizons. 
This is a byproduct of computing the velocity through finite difference of position.
The EqF with covariance intersection also shows large deviations, but they occur more smoothly as a result of the filtering process. 
The EqF with equivalent measurements produces low-variance velocity estimates and significantly outperforms the alternate algorithms.

%========================================================================================================%
\subsection{Image Tracking under Fast Motion} \label{subsec:results_fast}

\begin{figure}[htbp]
    \centering
    % Left Figure
    \begin{minipage}[b]{0.49\textwidth}
        \centering
        \includegraphics[width=\linewidth]{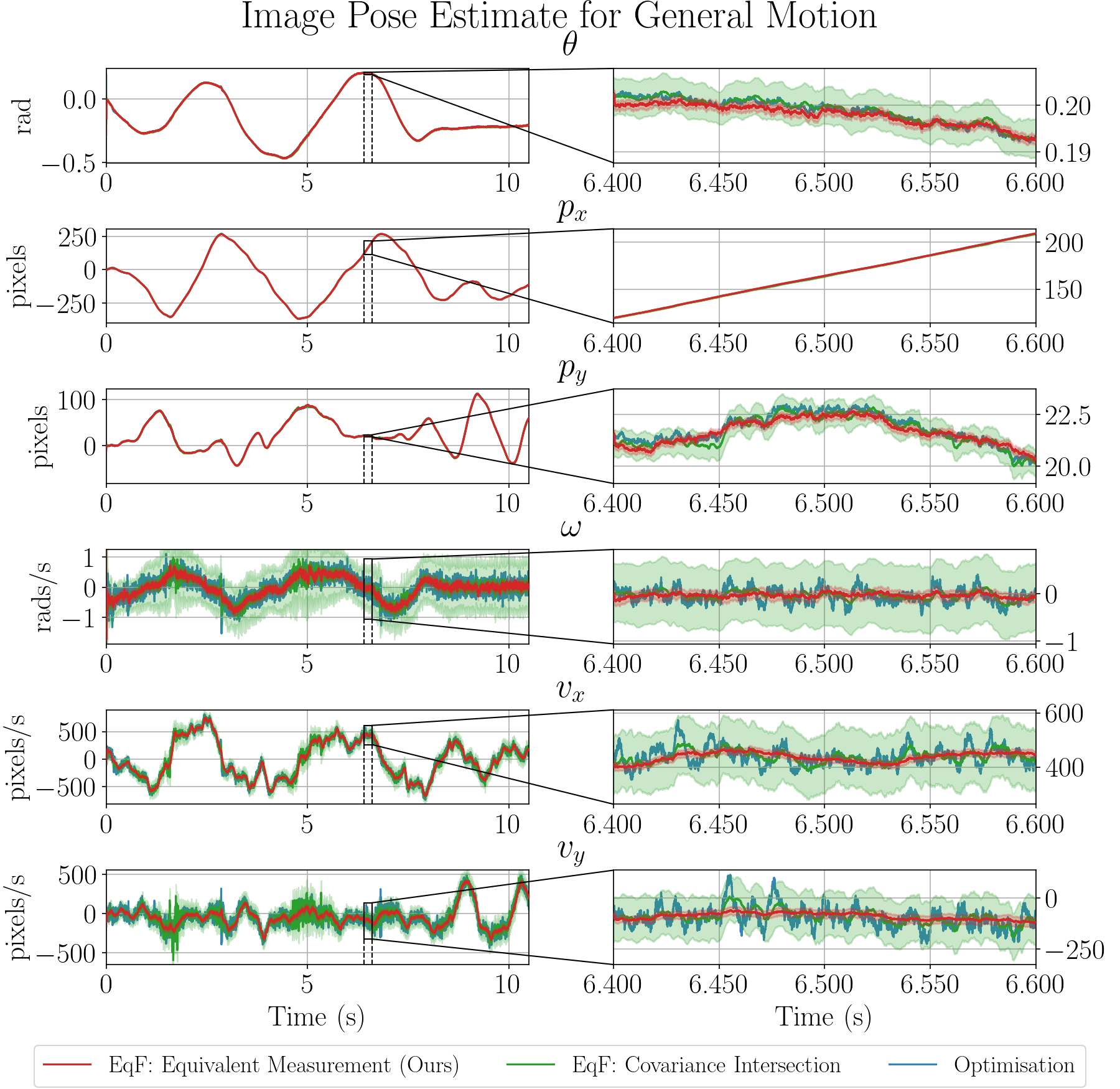}     
        \caption{State estimates (lines) and error covariance (one standard deviation, shaded) for general motion. The right column shows a zoomed time window indicated by the dashed lines in the left column.}
        \label{fig:genera_estimate_EM_OPT}
    \end{minipage}
    \hfill
    % Right Figure
    \begin{minipage}[b]{0.49\textwidth}
        \centering
        \includegraphics[width=\linewidth]{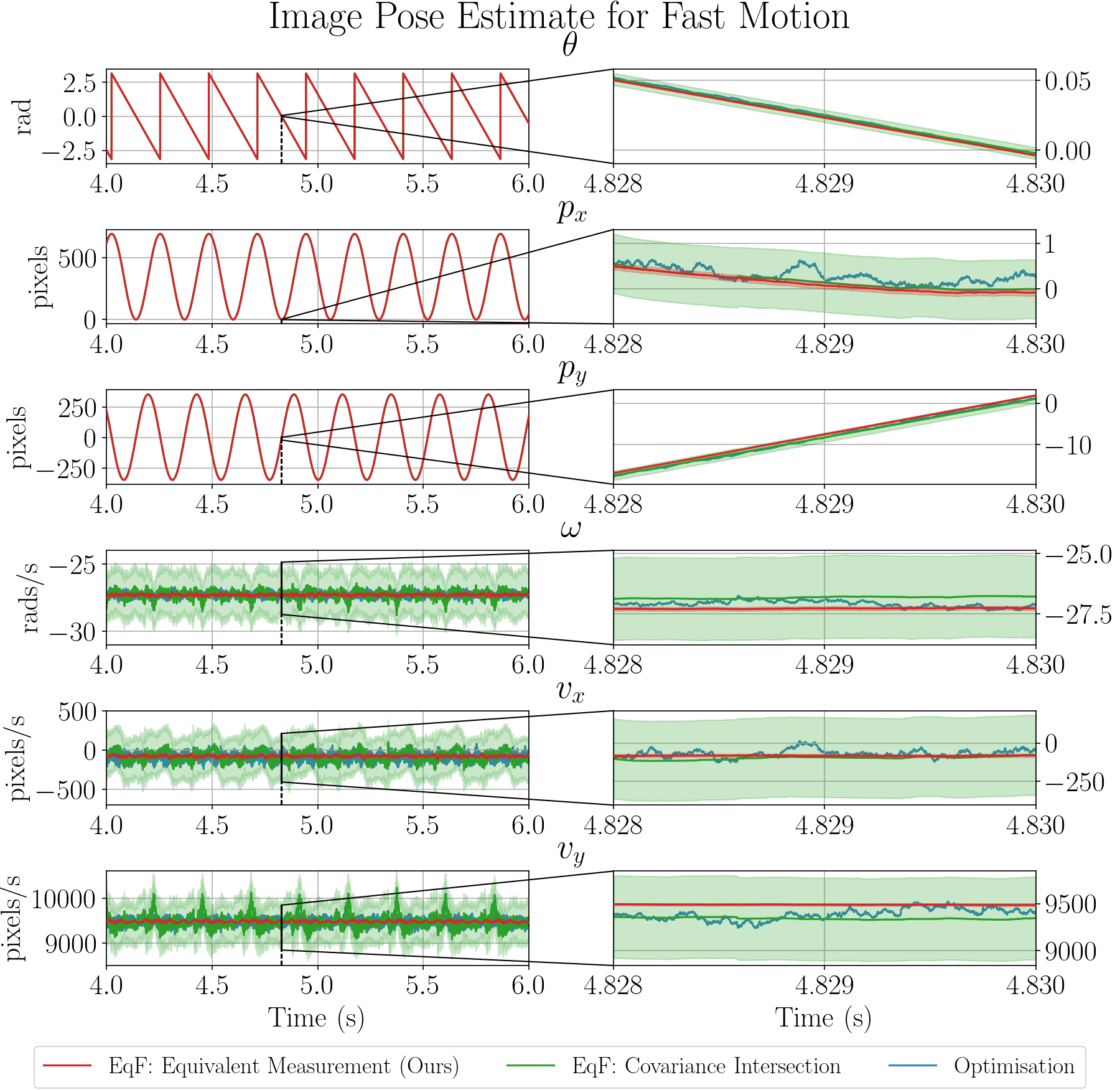}     
        \caption{State estimates (lines) and error covariance (one standard deviation, shaded) for fast motion. The right column shows a zoomed time window indicated by the dashed lines in the left column.}
        \label{fig:fast_estimate_EM_OPT}
    \end{minipage}
\end{figure}

We also compare the performance of the three algorithms for fast rotational motion (Fig.~\ref{fig:spinning_exp_setup}).
The state estimates are shown in Figure \ref{fig:fast_estimate_EM_OPT}.
Due to the fast motion, only a short time window is presented so that details can be resolved.
Similar to Section \ref{subsec:results_general}, all algorithms perform comparably at the macro-scale for $\torSE(2)$, and the $\se(2)$ estimates for optimisation and covariance intersection methods are visibly noisier than the equivalent measurement.
In the zoomed window, the equivalent measurement and covariance intersection perform comparably and provide smooth estimates over this short time interval.
As expected, covariance intersection stabilises to a significantly higher steady-state covariance.
For fast motion, the optimisation algorithm performs worse than the two filtering algorithms, with significantly more noise in both estimates. 
This can be attributed to the isolated estimation with no temporal correspondence, and faster motion inducing larger changes in feature tracker estimates that no longer satisfy the zero-order-hold assumption. 
This reinforces the advantages of filtering for image tracking as the temporal evolution of the system helps to stabilise the estimates. 

%========================================================================================================%
\subsection{Discussion} \label{subsec:discussion}

From Figures \ref{fig:genera_estimate_EM_OPT} and \ref{fig:fast_estimate_EM_OPT}, the EqF with equivalent measurement appears to outperform the two alternate approaches by providing smoother $\torSE(2)$ and $\se(2)$ estimates across both datasets. 
Although both alternate algorithms are able to perform image tracking, their estimates (particularly for $\se(2)$) are significantly noisier. 

First comparing to the EqF with covariance intersection. 
As a conservative estimator, covariance intersection is guaranteed to overestimate the covariance of the state for any cross-correlation which leads to overly cautious estimates.
Additionally, computing the weighting parameter $\alpha_t$ at each timestep (see Section \ref{subsec:comparisons}) introduces an optimisation problem that must be solved up to the event rate of the feature tracker.

The optimisation algorithm achieves comparable performance for $\torSE(2)$ in the presented datasets, but the velocity estimates are significantly worse.
The design-decisions required to compute the velocity directly impact the performance, whereas this is embedded in the filtering algorithms.
As there is no temporal dependence between estimates, it is more sensitive to occluded feature points, particularly when there are limited feature points.
In contrast the filtering-based approaches can use the kinematic model to overcome such situations and provide robustness against instances of temporary occlusions and lost features.

%====================================================%
%=====                 Conclusions              =====%
%====================================================%
\section{Conclusions} \label{sec:conclusions}

In this work, we propose an EqF for $\SE(2)$ image tracking with an event camera.
An event-based feature tracker provides sub-pixel position measurements at high temporal resolution, but this data is highly correlated as it is derived from a filtering process. 
We develop an equivalent measurement to handle the high temporal correlation between consecutive measurements.
We evaluate our algorithm experimentally on two datasets against two alternate approaches.
Our algorithm provides significantly smoother and more stable estimates than the alternate algorithms for features moving up to 7000 pixels per second. 
The equivalent measurement formulation is crucial in our filter, and with its development there is a clear path to extend our algorithm to $\SL(3)$ for full homography tracking.

%====================================================%
%=====                 References               =====%
%====================================================%

\bibliography{mybib}
\bibliographystyle{IEEEtran}
\end{document}